\documentclass{esannV2}
\usepackage[dvips]{graphicx}
\usepackage[latin1]{inputenc}
\usepackage{amssymb,amsmath,array}

\usepackage{todonotes}
\usepackage{subcaption}
\usepackage{hyperref}

\begin{document}
\title{Detecting Adversarial Examples through Nonlinear Dimensionality Reduction}

\author{Francesco Crecchi $^1$, Davide Bacciu $^1$, Battista Biggio$^2$
%
%
\vspace{.3cm}\\
%
1- Universit\'{a} di Pisa - Dipartimento di Informatica \\
 Largo Bruno Pontecorvo, 3, 56127 Pisa - Italy
%
\vspace{.1cm}\\
2-Universit\'{a} degli Studi di Cagliari - Dip. di Ingegneria Elettrica ed Elettronica \\
 Via Marengo, 09123 Cagliari - Italy \\
}

\maketitle

\begin{abstract}
Deep neural networks are vulnerable to adversarial examples, i.e., carefully-perturbed inputs aimed to mislead classification. This work proposes a detection method based on combining non-linear dimensionality reduction and density estimation techniques.
Our empirical findings show that the proposed approach is able to effectively detect adversarial examples crafted by non-adaptive attackers, i.e., not specifically tuned to bypass the detection method. Given our promising results, we plan to extend our analysis to adaptive attackers in future work.
\end{abstract}

\section{Introduction}
Deep neural networks (DNNs) reach state-of-the-art performances in a wide variety of pattern recognition tasks. However, similarly to other machine-learning algorithms~\cite{Biggio2013}, they are vulnerable to \textit{adversarial examples}, i.e., carefully-perturbed input samples that mislead classification~\cite{Szegedy2013}.

The existence of adversarial examples questions the use of deep learning solutions in safety-critical contexts, such as autonomous driving and medical diagnosis. Despite several approaches have been proposed to protect a classifier from adversarial examples, many of them turned out to be ineffective in front of \emph{adaptive} attacks that exploit knowledge of the defense mechanism~\cite{Carlini2017a, Athalye2018}. 
According to~\cite{Biggio2018}, we can categorize defense mechanisms that remain effective also against adaptive attackers into two main groups. The first ones include robust optimization techniques that can be exploited to reduce input sensitivity (or to increase the input margin) of the classifier; for example, \textit{adversarial training} obtains a smoother decision function by retraining the model on adversarial samples generated against it. This requires, however, generating the attacks and re-training the classifier on them, which is very computationally demanding for large models such as state-of-the-art convolutional neural networks, and it does not provide any performance guarantee for adversarial samples crafted via a different attack algorithm. 
The second line of effective defenses includes explicit detection and rejection strategies for adversarial samples. 

In recent years, different rejection-based countermeasures have been proposed. In~\cite{Feinman2017} it is proposed to distinguish natural samples from adversarial ones exploiting a Kernel Density Estimator on the embeddings obtained from the last hidden layer of the neural network. The idea of a layer-wise detector has been pushed forward in~\cite{Metzen2017} by providing multiple ``detector" subnetworks placed at different layers of the classifier which is intended to protect. Each subnetwork is trained to perform a binary classification task of distinguishing genuine data from samples containing adversarial perturbations. Such multilayer detectors have been showed to achieve relevant results in protecting against \textit{static} adversaries, i.e. those who only have access to the classification network but not to the detector, and to significantly hardener the task of producing adversarial samples for \textit{dynamic} adversaries, i.e. in which also the detector gradient is used by the attacker to craft adversarial samples. 

In this paper, we build on the core concepts of~\cite{Feinman2017, Metzen2017} by implementing a multilayer adversarial example detector based on t-SNE~\cite{VanDerMaaten2009}, a powerful manifold-learning technique mainly used for nonlinear dimensionality reduction. Our preliminary results are encouraging, showing that we can successfully detect adversarial examples crafted by non-adaptive attack algorithms. For this reason, we plan to further investigate and strengthen our method against adaptive attacks in the near future.

\section{Detection of Adversarial Examples using t-SNE}
High-dimensional data, such as images, are believed to lie in a low dimensional manifold embedded in a high dimensional space. In~\cite{Szegedy2013} it is speculated that adversarial examples come from ``pockets'' in the data manifold caused by the high non-linearity of deep networks. The adversarial inputs contained in such pockets occur with very low-probability which prevents them to be used during training and testing. Yet, these pockets are dense and so an adversarial example can be found near every test case but lies out-of-manifold.

In this paper, we support the so-called, \textit{manifold hypothesis} by empirically demonstrating that is it possible to detect adversarial samples exploiting a non-linear dimensionality reduction method such as t-SNE to separate in- and out-of-manifold samples.
\begin{figure}[t]
    \centering
        \begin{subfigure}[b]{0.49\textwidth}
            \includegraphics[width=\textwidth]{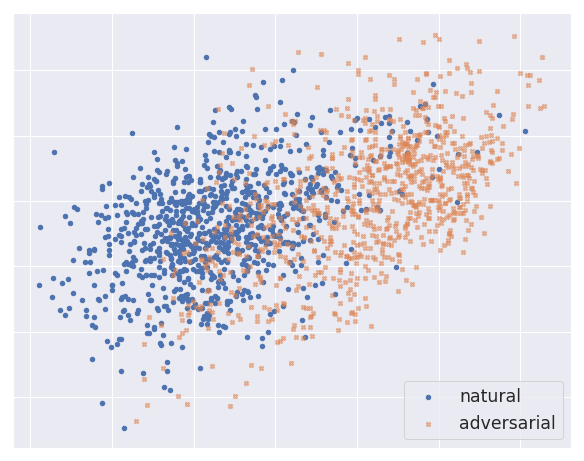}
        \end{subfigure}
        \hfill
        \begin{subfigure}[b]{0.49\textwidth}
            \includegraphics[width=\textwidth]{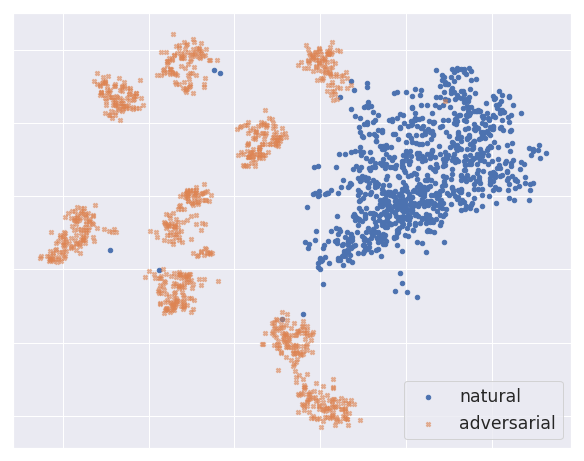}
        \end{subfigure}
        \caption{
        Separability test for the MNIST dataset: if t-SNE projections are learned from natural samples only, the learned map cannot separate natural from adversarial samples (left). Whereas, if mixing natural and adversarial samples when learning, the resulting map is able to separate them (right).
        }
        \label{fig:separability_test}
\end{figure}

Mostly used for data visualization, t-SNE maps high-dimensional data into low-dimensional spaces while maintaining the relative distances between samples. As such, we used this technique to investigate the separability of natural and adversarial samples in the projected space. As shown in Fig.\ref{fig:separability_test}, when the t-SNE projection is learned from a population of samples made of neural activations at a certain layer $l$ for both natural and adversarial network inputs then it can very well separate them into two different clusters. 

\begin{figure}[t]
    \centering
    \includegraphics[width=0.9\textwidth]{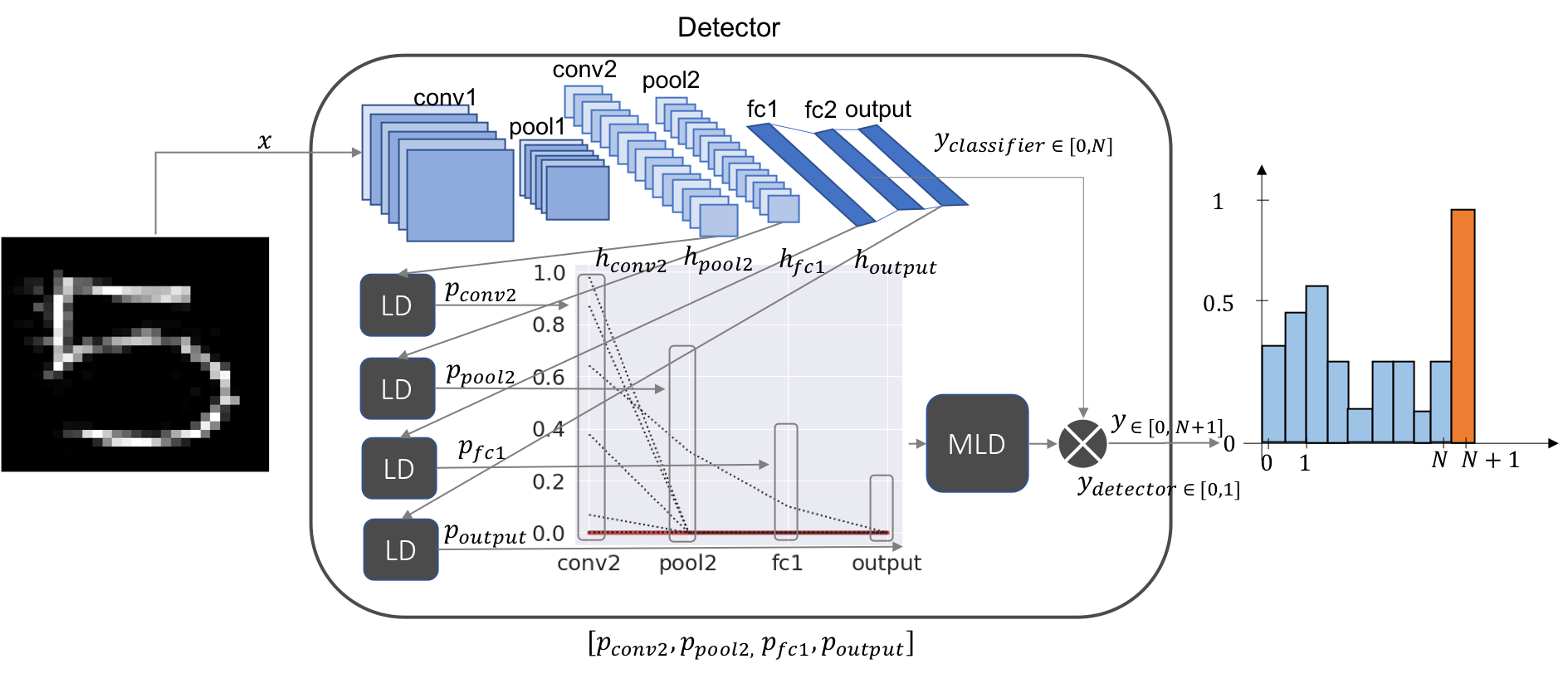}
    \caption{Detector architecture for a convolutional NN with four monitored layers.}
    \label{fig:detector}
\end{figure}

\vspace{2pt}
\noindent \textbf{Detection Procedure.} Figure~\ref{fig:detector} provides a representation of the detector architecture. Given an already-trained DNN classifier to be protected, we build on \cite{Metzen2017} by attaching a {\it Layer Detector} (LD) to potentially all classifier layers. Each LD takes as input $h_l(x)$, the internal representation of a given input sample $x$ extracted by the classifier at layer $l$ and outputs a vector of probability scores $p_l(x)$ for $x$ to be a natural sample of each of the target classes. The predictions of each LD form a data series that is fed into the $Multilayer Detector$ (MLD) which is responsible to capture trends that identify adversarial examples with respect to natural data. In case of detection of an adversarial attack ($y_{detector}=1$), the output of the classifier $y_{classifier} \in [0, N]$ (for $N$ target classes) is overridden by class $N+1$ to indicate that $x$ is an adversarial example. 

Attached to a given layer $l$ in the network architecture, $LD_l$ is composed by a \textit{Mapper} and an \textit{Estimator} for each target class. The \textit{Mapper} of a given class, exploits the t-SNE algorithm to map samples from the high-dimensional embedding space in which $h_l(x)$ lies, to a small-dimensional one (typically two or three dimensions) in which natural and adversarial samples for such class are \textit{likely} to be separated \footnote{Experiments show that the deeper in the classifier architecture is the layer which the LD is attached to, the better t-SNE is able to discriminate between natural and adversarial samples.}.
In this low dimensional space, a Kernel Density Estimator (KDE) is fitted on natural samples only, so to learn the distribution of natural data of a given target class.
When a new data sample $x'$ is to be classified and $h_l(x)$ is computed, the $LD_l$ (at layer $l$) use each class \textit{Mapper} to project this new point in a low dimensional space and then uses class \textit{Estimators} to score the probability of $x'$ to be of each target class. Intuitively $p_l(x)$, the probability-score vector of $LD_l$ for $x$, represents the likelihood of $x$ being from one of the target classes at layer $l$. 

In order to capture slight input perturbations, as in the case of adversarial examples, t-SNE has to be fed both with natural and adversarial examples (see the example in Fig.\ref{fig:separability_test}). As such, \textit{Mapper} training requires crafting adversarial samples \`{a} la \textit{Adversarial Training} \cite{Goodfellow2014}. Adversarial Training provides evidence of increasing robustness of the classifier with respect to the attack used for crafting training adversarial samples solely, nevertheless, it is widely accepted as the most reliable countermeasure provided that the attack used for training is general enough to subsume many different attack algorithms \cite{Madry2017}. The same assumption holds in our mapping algorithm: by crafting minimal-distance adversarial examples with a powerful attack as \cite{Carlini2017} during training, we found \textit{Mapper} to be able to separate adversarial samples also from other existing attack algorithms as \textit{Fast Gradient Sign Method} (FGSM)~\cite{Goodfellow2014}, \textit{Basic Iterative Gradient Sign Method} (BIM)~\cite{Kurakin2016} and \textit{Projected Gradient Descent} (PGD)~\cite{Madry2017}. A great drawback of Adversarial Training is that, since adversarial samples are used to regularize the model during training, they have to be generated at each epoch. This is very computationally demanding and needs \textit{fast} algorithms to craft adversarial samples (e.g., FGSM), which produces sub-optimal solutions for the attacker optimization problem. Our detector, instead, requires to generate adversarial samples only \textit{once} before starting to train, as the very same adversarial samples are used to train the LDs altogether.

\begin{figure}[t]
    \centering
        \begin{subfigure}[b]{0.49\textwidth}
            \includegraphics[width=\textwidth]{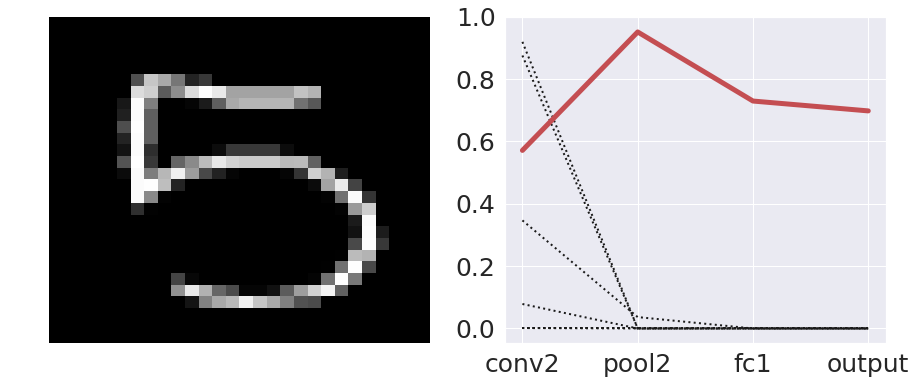}
        \end{subfigure}
        \begin{subfigure}[b]{0.49\textwidth}
            \includegraphics[width=\textwidth]{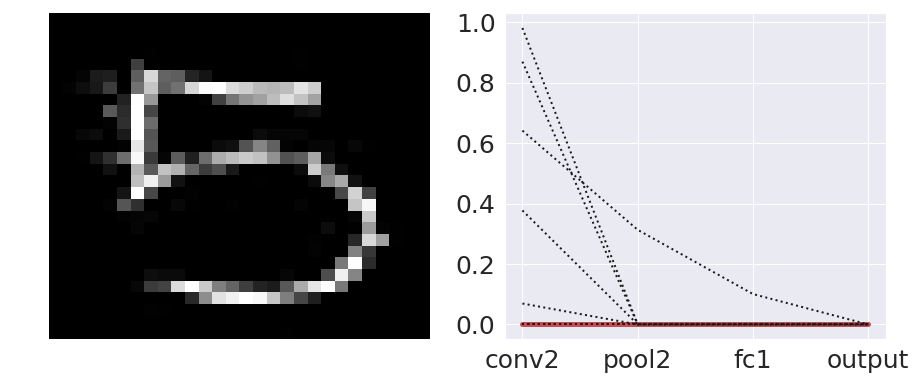}
        \end{subfigure}
        \caption{Comparison of $p$ scores data series for a natural sample from MNIST dataset (left) and a corresponding adversarial example (right). Solid line identifies the $p$ scores data series of the original target label for the input sample, whereas dashed lines are for other classes.
        }
        \label{fig:mnist_pscores_seq}
\end{figure}

The rationale for treating the output of each LD as a whole comes from observing the trends of the $p$ scores across layers for natural and adversarial samples. In Fig. \ref{fig:mnist_pscores_seq}, it is reported a comparison between the series of $p$ scores produced for a given natural example (on the left) and a corresponding adversarial one (on the right).  The two data series present significative differences in trends: as one may expect, in case of natural data the series of $p$ scores across layers achieve high values for the target class (i.e., above $0.5$) consistently (see Fig. \ref{fig:mnist_pscores_seq} left). Whereas, for adversarial samples, the $p$ scores series for the target class drops to values slightly above zero, mixing with the data series of the non-target classes, resulting in misclassification of the sample.


\section{Experiments}
\begin{figure}[t]
    \centering
        \begin{subfigure}[b]{0.2425\textwidth}
            \includegraphics[width=\textwidth]{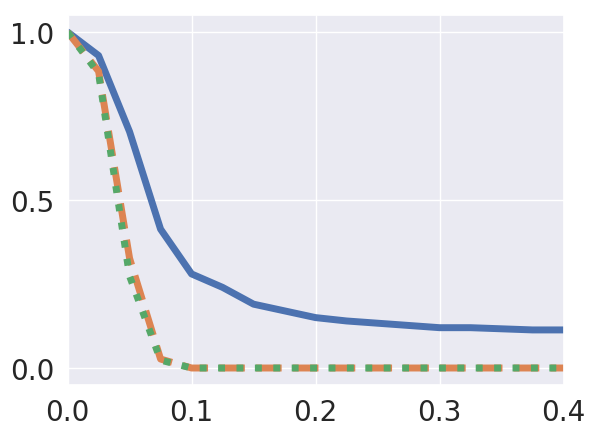}
        \end{subfigure}
        \begin{subfigure}[b]{0.2425\textwidth}
            \includegraphics[width=\textwidth]{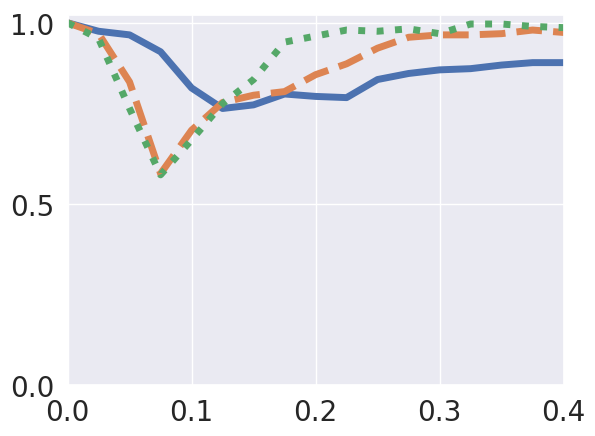}
        \end{subfigure}
        \begin{subfigure}[b]{0.2425\textwidth}
            \includegraphics[width=\textwidth]{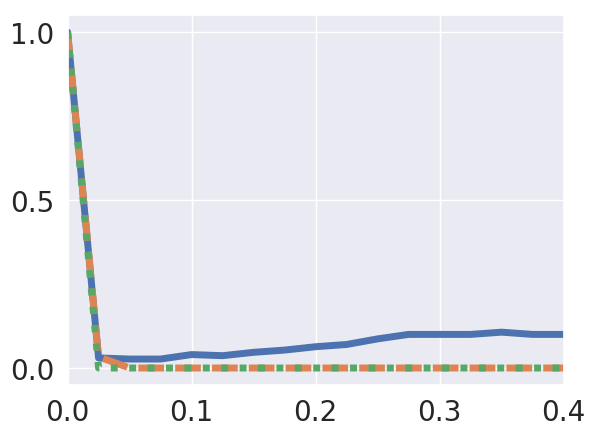}
        \end{subfigure}
        \begin{subfigure}[b]{0.2425\textwidth}
            \includegraphics[width=\textwidth]{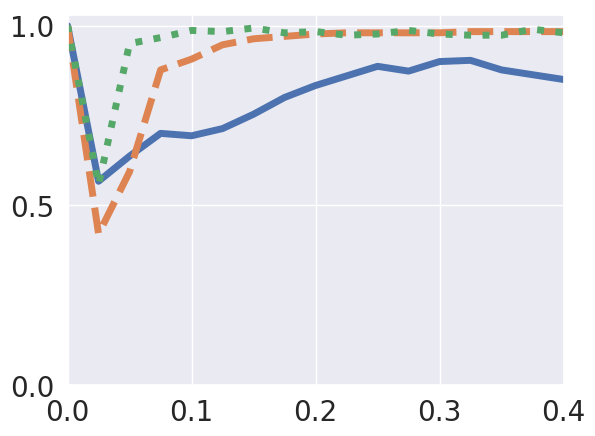}
        \end{subfigure}
        \caption{Security evaluation curves for classifier and detector on MNIST and CIFAR10 datasets, respectively. We used solid lines for FGSM attack, dashed lines for BIM and dotted lines for PGD. }
       \label{fig:sec_curves}
\end{figure}

Experiments assess the detection capabilities of the proposed detector in protecting a convolutional neural network (CNN) of \textit{LeNet}~\cite{LeCun1998a} type for image recognition on MNIST and CIFAR10 datasets. The accuracy of the classifier on test data drops from $0.98$ for MNIST dataset and from $0.77$ for CIFAR10 to approaching zero when samples are perturbed with adversarial noise. Wrapping the classifier with our detector allows maintaining high accuracy on slightly perturbed samples (i.e. on which the perturbation does not induce the label change) while detecting adversarially perturbed ones (i.e. in which the label change is induced). In our experiments, we crafted the adversarial samples needed to fit the \textit{Mappers} using Carlini-Wagner attack~\cite{Carlini2017}.
Protecting an $N$ class classifier from adversarial examples is cast as a $N+1$ classification problem, where the $N+1$ class represents adversarial samples. We measured the increase in robustness against adversarial examples by means of \textit{security evaluation curves} \cite{Biggio2014}. Given a classifier to be evaluated, this amounts to testing its accuracy (on y-axis in the figures) with respect to increasing adversarial perturbations (x-axis). We performed a security evaluation of the CNN classifier and of our detector against state-of-the-art adversarial attacks such as FGSM, BIM, and PGD using $L_2$ norm as distance metrics between samples. In Fig. \ref{fig:sec_curves} are reported the security curves constructed for the classifier the detector respectively for MNIST and CIFAR10 dataset. It can be noticed that, by exploiting $p$ values trends, our detector is able to protect the CNN classifier ``under attack'' by maintaining the accuracy above $0.8$ for MNIST and above $0.7$ for CIFAR10, on average. 

\section {Contributions, Limitations and Future Work}
We introduced a new detector for adversarial examples which exploits manifold learning techniques to identify adversarial samples from natural ones. Our findings provide experimental support to the \textit{manifold hypothesis} which identifies adversarial examples as out-of-distribution samples in the neural representation manifolds. This renews the hope to be able to protect an existing classifier by wrapping it with a detector that identifies and reject out-of-distribution data. In future work, we intend to evaluate our detector against adaptive attacks that also try to bypass the detection method and to compare it with other existing detection techniques.

\vspace{5pt}
\noindent \textbf{Acknowledgments.} The authors would like to thank Daniele Tantari, Scuola Normale Superiore, Pisa, Italy, for providing useful feedback on the definition of our statistical approach.


\begin{footnotesize}




\bibliographystyle{unsrt}
\bibliography{references}

\end{footnotesize}


\end{document}